\documentclass[conference]{ieeeconf}  

\IEEEoverridecommandlockouts                        
\title{\LARGE \bf
  
    NLiPsCalib: An Efficient Calibration Framework for High-Fidelity 3D Reconstruction of Curved Visuotactile Sensors

}

\author{
    Xuhao Qin$^{*}$,
    Feiyu Zhao$^{*}$,
    Yatao Leng,
    Runze Hu, and
    Chenxi Xiao\textsuperscript{\dag}
    \thanks{$^{*}$ Equal contribution.}
    \thanks{\textsuperscript{\dag} Corresponding author.}
    \thanks{
        This work was supported by the Natural Science Foundation of Shanghai (Grant No. 25ZR1402370), and partially by Shanghai Frontiers Science Center of Human-centered Artificial Intelligence (ShangHAI), MoE Key Laboratory of Intelligent Perception and Human-Machine Collaboration (KLIP-HuMaCo). The experiments were supported by the Core Facility Platform of Computer Science and Communication, SIST, ShanghaiTech University.
    } 
    \thanks{
        Xuhao Qin, Feiyu Zhao, Yatao Leng, Runze Hu, and Chenxi Xiao 
        are with the School of Information Science and Technology,
        ShanghaiTech University.
        {\tt\footnotesize \{qinxh2024, zhaofy12024, lyt, hurz2024, xiaochx\}@shanghaitech.edu.cn}
    }
    \thanks{
        Project webpage: \href{https://nlipscalib.github.io/}{https://nlipscalib.github.io/}
    }
}

\usepackage{adjustbox} 
\usepackage{balance}
\usepackage{algorithm}
\usepackage[utf8]{inputenc}
\usepackage{newunicodechar}
\newunicodechar{，}{,}
\usepackage{bbm}
\usepackage{etoolbox}
\usepackage{multicol}
\usepackage{cite}
\usepackage{graphics} 
\usepackage{times} 
\usepackage{amsmath} 
\usepackage{amssymb} 

\usepackage{svg}
\usepackage{mathrsfs}
\usepackage{amsfonts}
\usepackage{wrapfig}
\usepackage{amsmath}

\usepackage{kantlipsum}
\usepackage{subcaption}
\usepackage{marvosym}

\usepackage{float}
\usepackage{ctable}
\usepackage{cuted}
\usepackage{colortbl}
\usepackage{multirow}
\usepackage{wrapfig}
\usepackage[misc,geometry]{ifsym}
\usepackage{algorithm}
\usepackage{algpseudocode}
\usepackage{xcolor}   
\usepackage{pifont}    
\usepackage{graphicx}
\usepackage{caption}
\usepackage{subcaption}

\usepackage{arydshln}

\usepackage{array}
\usepackage{bm}
\usepackage{caption}
\usepackage{setspace}
\let\labelindent\relax
\usepackage{enumitem}
\usepackage{threeparttable}
\usepackage{makecell}  
\usepackage{booktabs,array}
\usepackage{float}

\newcolumntype{L}[1]{>{\centering\arraybackslash}m{#1}}

\newcommand{\TODO}[1][]{\textcolor{red}{\bf [TODO]}}
\setlist[enumerate,1]{itemsep=3pt}

\definecolor{formalgreen}{rgb}{0.1, 0.7, 0.1} 
\definecolor{formalred}{rgb}{0.9, 0.2, 0.2}

\makeatletter
\let\NAT@parse\undefined
\makeatother
\usepackage{hyperref} 
\hypersetup{
  colorlinks   = true,   
  linkcolor    = red,    
  citecolor    = blue,    
}

\begin{document}

\maketitle
\thispagestyle{empty}
\pagestyle{empty}

{
\begin{abstract}
Recent advances in visuotactile sensors increasingly employ biomimetic curved surfaces to enhance sensorimotor capabilities. Although such curved visuotactile sensors enable more conformal object contact, their perceptual quality is often degraded by non-uniform illumination, which reduces reconstruction accuracy and typically necessitates calibration. Existing calibration methods commonly rely on customized indenters and specialized devices to collect large-scale photometric data, but these processes are expensive and labor-intensive. To overcome these calibration challenges, we present \textbf{NLiPsCalib}, a physics-consistent and efficient calibration framework for curved visuotactile sensors. NLiPsCalib integrates controllable near-field light sources and leverages Near-Light Photometric Stereo (NLiPs) to estimate contact geometry, simplifying calibration to just a few simple contacts with everyday objects. We further introduce \textbf{NLiPsTac}, a controllable-light-source tactile sensor developed to validate our framework. Experimental results demonstrate that our approach enables high-fidelity 3D reconstruction across diverse curved form factors with a simple calibration procedure. We emphasize that our approach lowers the barrier to developing customized visuotactile sensors of diverse geometries, thereby making visuotactile sensing more accessible to the broader community. 
\end{abstract}
}

\section{Introduction}

Robots often adopt biomimetic or functional designs in both morphology and sensory systems\cite{WANG2021100001}. In line with this trend, robotic tactile sensors have evolved beyond conventional flat architectures to encompass diverse form factors. For example, in humanoid robots, curved and rounded fingertip-inspired sensors are now widely used in dexterous robotic hands, as they mimic the geometry of fingertips and finger pads, enabling stable, conformal contact with objects and supporting omnidirectional perception~\cite{ward2018tactip, sun2022soft, piacenza2020sensorized, padmanabha2020omnitact}. Beyond humanoid robots, application-specific tactile sensors are also emerging, including cylindrical designs for industrial arms~\cite{zhang2024gelroller} and slim-profile sensors for minimally invasive robotic surgery~\cite{yu2024orbit,minitac}. Thus, the capability to design and fabricate tactile sensors in customized shapes is essential for broadening the scope of tactile perception and manipulation.

\begin{figure}[t] \centering
    \includegraphics[width=0.98\linewidth]{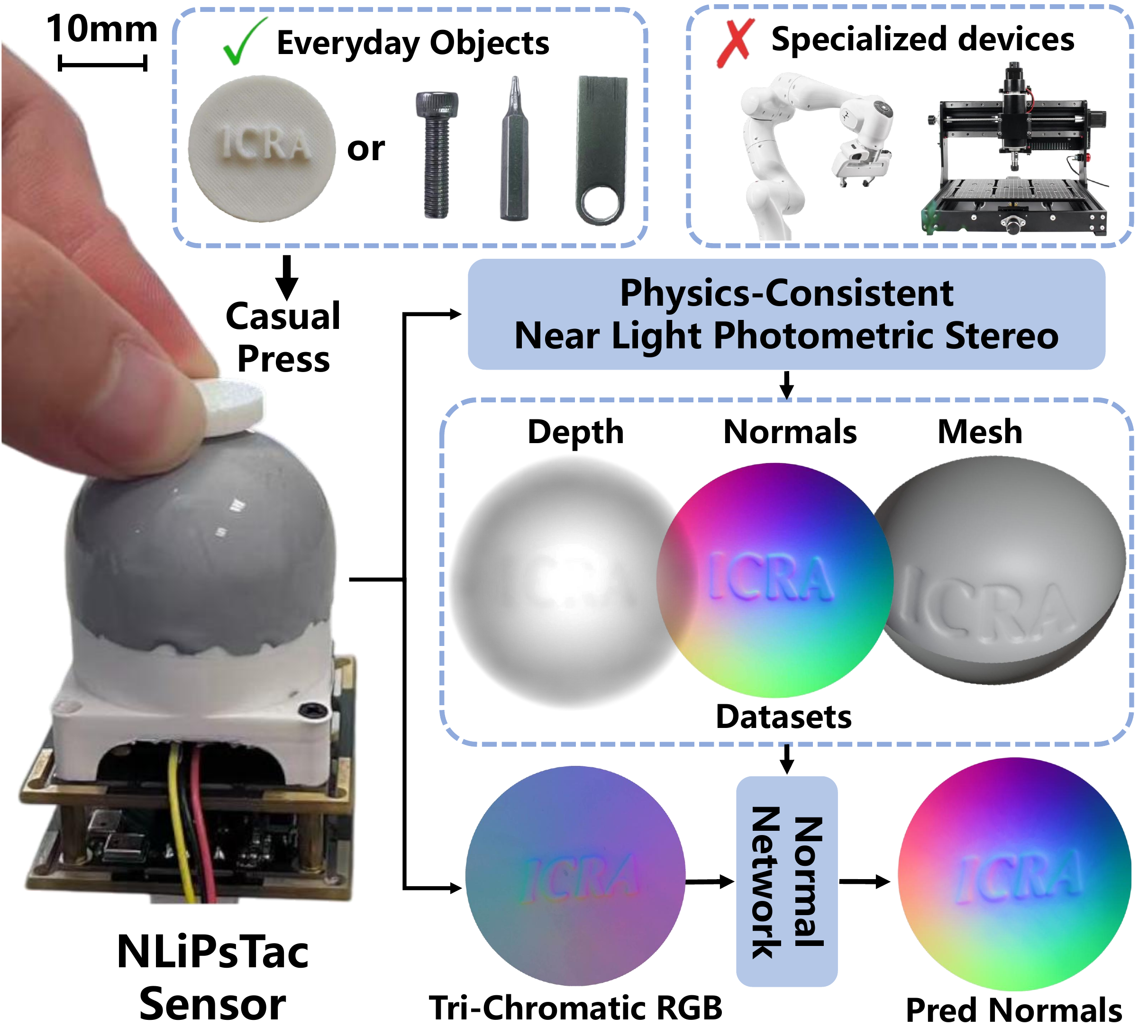}
    \vspace{-5pt}
    \caption{NLiPsCalib, a new calibration pipeline for curved visuotactile sensors that requires no specialized devices. 
    It leverages casual presses with everyday objects, followed by near-light photometric stereo to obtain accurate geometries for building the calibration dataset. 
    This dataset further enables training a neural network for real-time normal inference.}
    \label{fig:teaser}
    \vspace{-3mm}
\end{figure}

Among these technologies, visuotactile sensors have gained prominence due to their high-resolution 3D shape sensing capabilities and inherently customizable form factors \cite{9028163}. Over the past decade, numerous designs have been introduced and deployed on robotic manipulators. Representative examples range from flat sensors like GelSight \cite{yuan2017gelsight}, 9DTact \cite{9dtact}, and DelTact \cite{Deltact} to curved, finger-shaped sensors such as the Soft Finger Sensor \cite{romero2020soft} and GelSight Svelte \cite{zhao2023gelsight}.

Despite this progress, reliably extracting high-fidelity depth information from a sensor’s output remains a persistent challenge. The standard pipeline for this task leverages photometric stereo: first, illumination patterns are mapped to surface normals; then, the normals are numerically integrated to reconstruct the 3D shape \cite{johnson2009retrographic,yuan2017gelsight}. 
A critical bottleneck of this approach lies in establishing an accurate mapping from illumination to normals. This mapping is complex because the internal illumination is inherently non-uniform: light intensity varies across the surface due to the elastomer’s curvature and near-field effects from embedded light sources, which cause distance-dependent attenuation.

To model these compound lighting effects, existing methods typically employ data-driven approaches that learn the mapping from collected intensity images to ground-truth surface normals. However, acquiring this ground-truth data is non-trivial. It requires either constructing a high-fidelity digital twin of the sensor for simulation \cite{gomes2023beyond,lin20253d,agarwal2025vision} or using precisely calibrated hardware such as CNC-machined probes or professional indentation devices \cite{tippur2023gelsight360}. Both approaches are costly in time, effort, and resources, directly undermining the goal of rapid and accessible customization for visuotactile sensors.

In this paper, we introduce NLiPsCalib, a novel calibration technique for visuotactile sensors that eliminates the need for specialized equipment (Fig.~\ref{fig:teaser}). Our key insight is that the physics-based Near-Light Photometric Stereo (NLiPs) model \cite{NLiPs} is highly consistent with the internal illumination conditions of curved visuotactile sensors. This enables high-quality shape calibration using only the sensor’s multiple internal light sources.
The calibration process is simple: it requires only a few physical interactions, such as pressing the sensor against textured everyday objects. For each pressed indentation, NLiPsCalib performs a single run of Near-Light Photometric Stereo to capture the resulting image data, from which it directly estimates the indented geometry and surface normals without requiring the ground-truth shape of the indenting object, thereby significantly simplifying the creation of a calibration dataset.

We evaluate the NLiPsCalib framework on a physical sensor. To this end, we introduce NLiPsTac, a modular tactile sensor featuring individually controllable light sources and an adaptable structure for various elastomer form factors. 
Performing calibration on this sensor requires around 50 casual presses on everyday objects, and the resultant calibration quality is comparable to state-of-the-art device-based methods in terms of reconstruction error \cite{yuan2017gelsight, zhang2024gelroller,tippur2024rainbowsight}. Furthermore, we validate our approach on elastomers of different shapes to demonstrate its adaptability. By significantly reducing the effort and expertise required for calibration, our work makes the design and deployment of custom visuotactile sensors more practical and accessible.

In summary, our main contributions are:
\begin{enumerate}
    \item \textbf{NLiPsCalib}: A physics-based and data-efficient calibration framework for curved visuotactile sensors that requires no specialized external hardware.
    \item The adaptation and validation of the Near-Light Photometric Stereo (NLiPs) model for 3D shape reconstruction in tactile sensing.
    \item \textbf{NLiPsTac}: A hardware platform designed for developing and testing near-light visuotactile sensors.
    \item A comprehensive evaluation of the proposed technique, along with discussions and insights for promoting broader usage.
\end{enumerate}

\vspace{-2mm}

\section{Related Works}

\begin{figure*}[t]
    \centering
    \includegraphics[width=\textwidth]{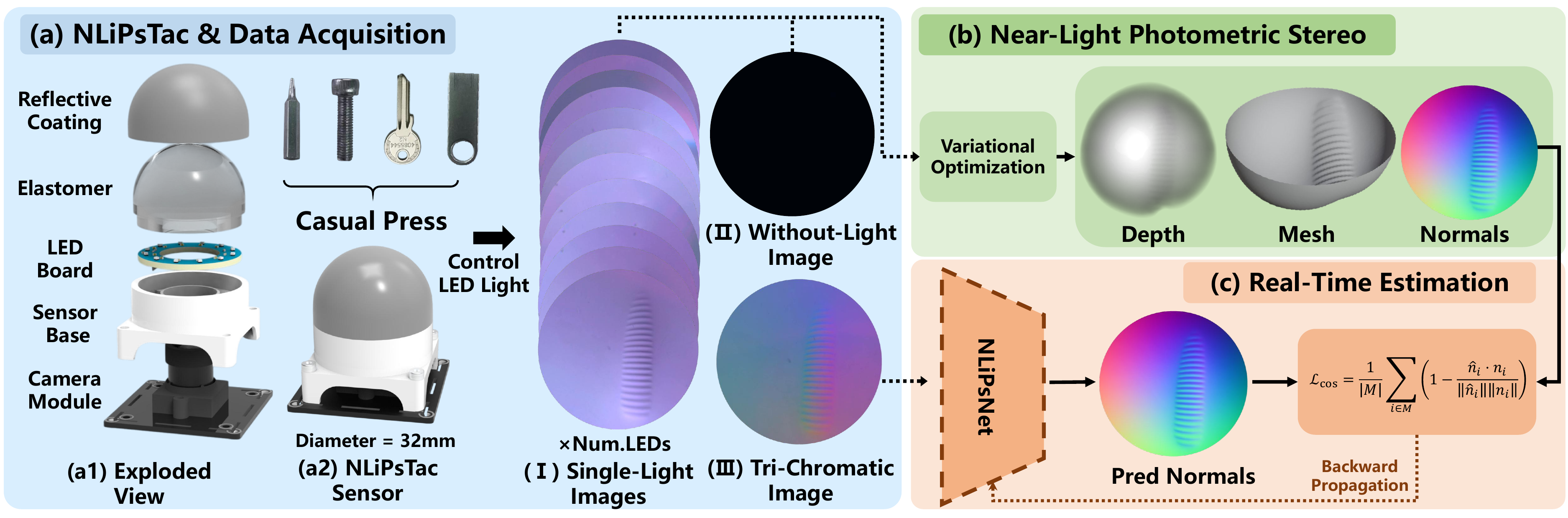}
    \caption{System pipeline. Using the proposed NLiPsTac tactile sensor, the framework collects a calibration dataset with NLiPs, enabling the training of NLiPsNet, a network designed for real-time 3D shape inference under trichromatic illumination.}
    \label{fig:pipeline}
    \vspace{-6mm}
\end{figure*}

\subsection{Visuotactile Sensors}
Tactile sensing is essential for robotic manipulation, as it provides contact information that vision alone cannot capture. Among various designs, visuotactile sensors have gained particular attention for their ability to deliver high-resolution contact geometry. Early work mainly focused on planar sensors, such as GelSight~\cite{yuan2017gelsight}, 9DTact~\cite{9dtact}, and DelTact~\cite{Deltact}, which demonstrated strong performance in dexterous manipulation tasks. However, planar structures are most suitable for parallel grippers and are less compatible with biomimetic mechanisms such as humanoid fingertips~\cite{kim2014roboray}, arms~\cite{kim2017anthropomorphic}, or other curved morphologies.  

More recently, curved elastomer geometries have been increasingly adopted to improve contact coverage and stability. Examples include the Soft Finger Sensor~\cite{romero2020soft}, OmniTact~\cite{padmanabha2020omnitact}, GelSight Svelte~\cite{zhao2023gelsight}, GelStereo Tip\cite{GelStereoTip} and R-Tac~\cite{lin2025pp}, which feature fingertip-inspired designs that enable stable, multi-directional contact. Building on this trend, we introduce \textbf{NLiPsTac}, a curved visuotactile sensor inspired by the human fingertip, designed to provide omnidirectional contact perception.

\begin{table}[t]
\centering
\resizebox{1.0\columnwidth}{!}{
\begin{threeparttable}
\caption{Methods and devices for calibrating curved visuotactile sensors often require external machines or specialized parts, whereas our sensor achieves calibration without them.}
\label{tab:CalibrationDevices}
\renewcommand{\arraystretch}{1.0}
\setlength{\tabcolsep}{5pt}
\begin{tabular}{cc}
\hline\hline
\textbf{Sensor} & \textbf{Calibration Method or Device} \\ \hline
GelSight360~\cite{tippur2023gelsight360}   & CNC machine indentation \\
RainbowSight~\cite{tippur2024rainbowsight} & CNC machine indentation \\
DenseTact 2.0~\cite{do2023densetact}       & CNC machine indentation \\
GelSplitter3D~\cite{lin20253d}             & Robotic arm indentation \\
HumanFT~\cite{wu2025humanft}               & 3D-printed calibration rig \\
R-Tac~\cite{lin2025pp}                     & 3D-printed calibration rig \\ 
\textbf{NLiPsTac (Ours)}                   & \textbf{NLiPs + Everyday objects} \\ \hline\hline
\end{tabular}
\end{threeparttable}
}
\vspace{-0.95em}
\end{table}

\vspace{-2mm}

\subsection{Reconstruction and Calibration for Visuotactile Sensing}
Accurate 3D reconstruction of contact geometry is fundamental to a wide range of tactile perception tasks, empowering the estimation of object's shape~\cite{Tac2Structure,huang2025gelslam}, pose~\cite{Tac2Pose,suresh2024neuralfeels}, and texture~\cite{li2013sensing}. Traditionally, reconstruction has been performed using different principles. For example, GelSight employs photometric stereo, while DTact relies on light intensity under single-light illumination~\cite{lin2022dtact}. However, these methods depend on mappings between light intensity and either surface shape or normals, and the quality of such mappings directly affects reconstruction accuracy. Moreover, while mappings can be reliably established on flat surfaces, curved-surface sensors inherently suffer from uneven illumination and near-field effects~\cite{ji2022model}, which invalidate parallel-light assumptions and compromise reconstruction stability.

To improve mapping fidelity, most existing approaches perform sensor calibration prior to reconstruction. For example, mechanical indentation is a widely used method \cite{yuan2017gelsight,tippur2023gelsight360,tippur2024rainbowsight,lin20253d}, typically involving a standard ball probe driven by a CNC machine or a robotic platform. Some studies have also employed customized 3D-printed parts for controlled indentation \cite{do2023densetact,wu2025humanft,lin2025pp}. Across these works, the key requirement is knowledge of the indenter’s pose and geometry, which makes the ground-truth surface normals accessible and thus enables pairing with RGB observations. Although effective, these calibration procedures are often complex, requiring specialized hardware (e.g., CNC machines or spherical probes) or extensive datasets, which increases cost and limits scalability.

To address these challenges, we propose \textbf{NLiPsCalib}, a calibration framework based on Near-Light Photometric Stereo (NLiPs)~\cite{NLiPs}. By explicitly modeling attenuation and the spatial distribution of near-field illumination, NLiPsCalib enables efficient and high-quality reconstruction while greatly simplifying calibration, thereby reducing reliance on specialized hardware and large datasets. This approach facilitates fast, low-cost, and high-fidelity tactile perception in curved-surface visuotactile sensors. Comparison between NLiPsCalib and previous works is shown in Tab.~\ref{tab:CalibrationDevices}.

\section{Overview\label{sec:overview}}

The goal of calibration is to establish the mapping between light intensities and surface geometry, serving as a prerequisite for high-fidelity 3D reconstruction. Specifically, the calibration process aims to obtain the mapping function $f$ between pixel intensities and surface normals, expressed as:  
\begin{equation}
\mathbf{n} = f(u, v, r, g, b),
\label{eq:normal_principle}
\end{equation}
where $(u,v)$ denote pixel coordinates and $(r,g,b)$ represent the corresponding light intensities in R, G, B color channels, respectively.  

A key challenge in calibration lies in obtaining accurate geometry $\mathbf{n}$ at the indented location, enabling the derivation of function $f$ for normal inference during test time.

To address this challenge, we introduce \textbf{NLiPsCalib}, a unified sensor calibration framework that derives ground-truth geometries directly from multi-source photometric cues. The design goal of \textbf{NLiPsCalib} is to infer surface normals $\mathbf{n}$ directly using near-light photometric stereo (Sec.~\ref{sec:calibration_method}). This approach offers two key advantages. First, it enables direct estimation without requiring costly equipment such as CNC-machined probes or robotic arms. Second, it avoids spatial misalignment between ground-truth normals $\mathbf{n}$ and intensity measurements $(u, v, r, g, b)$, thereby enabling the creation of high-fidelity paired calibration datasets for learning $f$. Based on the dataset, we further present \textbf{NLiPsNet}, a neural network for real-time normal prediction during deployment (Sec.~\ref{sec:real_time_normal_network}). 

Finally, we describe the design of a novel tactile sensor that serves as a testbed for validating the proposed framework (Sec.~\ref{sec:Sensor Design}). The overall system pipeline is illustrated in Fig.~\ref{fig:pipeline}.


\section{Sensor Calibration For Normal Estimation}
\label{sec:Sensor Calibration}

\subsection{Geometry Calibration with Near-light Photometric Stereo}\label{sec:calibration_method}
Our calibration pipeline generates high-fidelity surface geometry directly from sensor images, eliminating the need for external measurement hardware. To achieve this, we employ NLiPs framework, which is originally developed for high fidelity 3D reconstruction tasks such as 3D scanning for surface micro-textures \cite{NLiPs}. In this work, we adapt NLiPs to the optical environment of visuotactile sensors. The core contribution lies in its novel use for generating high-fidelity ground-truth data without external devices. In contrast to parallel illumination commonly used in previous works \cite{yuan2017gelsight}, NLiPs is more suitable for non-uniform illumination, thus naturally fits curved surfaces.

The goal of NLiPs is to compute a dense, pixel-wise depth map, $z(\mathbf{p})$, of the deformed surface. From this depth map, we can then derive the surface normals, estimate the surface albedo, and directly reconstruct the 3D surface points as a dense point cloud. The process involves two key components: a physical model of light intensity and an optimization procedure to estimate the depth map.

\paragraph{The Physical Model of Light Intensity}
The NLiPs model is built on a physical description of how light from a known point source reflects off an unknown surface and is captured by the camera. We define the sensor surface in the camera's coordinate system. For each pixel $\mathbf{p} = (u_p, v_p)$, the corresponding 3D point on the surface is $\mathbf{x}(\mathbf{p}) = (u, v, z(\mathbf{p}))$. Here, $(u,v)$ are the pixel coordinates converted to metric units using the camera's intrinsic parameters, and $z(\mathbf{p})$ is the unknown depth value we aim to recover.

For the $i$-th LED, located at a pre-calibrated 3D position $\mathbf{x}^i_s$, the intensity $I_{i,c}(\mathbf{p})$ observed at a pixel $\mathbf{p}$ in color channel $c \in \{R,G,B\}$ is given by:
\begin{align}
I_{i,c}(\mathbf{p}) 
&= \Psi_{i,c}\,\rho_{c}(\mathbf{p})
   \left[
      \frac{\mathbf{n}^i_s \cdot 
            \big(\mathbf{x}(\mathbf{p}) - \mathbf{x}^i_s\big)}
           {\big\|\mathbf{x}(\mathbf{p}) - \mathbf{x}^i_s\big\|}
   \right]^{\mu^i} \notag \\
&\quad \times 
   \frac{\big\{\big(\mathbf{x}^i_s - \mathbf{x}(\mathbf{p})\big) 
           \cdot \mathbf{n}(\mathbf{p})\big\}_+}
        {\big\|\mathbf{x}^i_s - \mathbf{x}(\mathbf{p})\big\|^3}
\label{eq:intensity_model}
\end{align}

In this model, the unknowns are the surface geometry and its reflectance. The geometry is described by the 3D positions $\mathbf{x}(\mathbf{p})$ and the unit surface normals $\mathbf{n}(\mathbf{p})$, while reflectance is described by the albedo $\rho_c(\mathbf{p})$. Note that both $\mathbf{x}(\mathbf{p})$ and $\mathbf{n}(\mathbf{p})$ are fundamentally determined by the depth map $z(\mathbf{p})$. The remaining terms are known parameters of the light sources: $\Psi_{i,c}$ is the calibrated intensity, $\mathbf{n}^i_s$ is the principal direction (Fig.~\ref{fig:Lighting Modeling}), and $\mu^i$ is the anisotropy parameter. The operator $\{\cdot\}_+ = \max(\cdot, 0)$ accounts for self-shadowing.

\paragraph{Calculating the Depth Map}
With the intensity model in place, the task is to recover the surface geometry (represented by the depth map $z(\mathbf{p})$) from multiple intensity images captured under different lighting conditions. In our case, these conditions are generated by sequentially activating the LEDs inside the sensor (refer to Sec.~\ref{sec:Sensor Design}). Each lighting condition provides data for solving Eq.~(\ref{eq:intensity_model}).

Based on captured data, directly solving Eq.~(\ref{eq:intensity_model})  
for $z(\mathbf{p})$ is difficult because the problem is highly non-linear, mainly due to 
the depth-dependent distance terms in the denominator. In addition, it is not possible to solve the surface 
normal $\mathbf{n}(\mathbf{p})$ independent of the depth $z(\mathbf{p})$, since this breaks the 
geometric constraint that the normal is the gradient of the surface, which may result in 
inconsistent results.

To address these challenges, we adopt the variational optimization framework from \cite{NLiPs}. This framework resolves this by framing the optimization solely in terms of the depth map $z(\mathbf{p})$. In other words, the surface normal $\mathbf{n}(\mathbf{p})$ is not an independent unknown. Instead, it is explicitly computed as a function of the depth map's spatial derivatives at each iteration. For a surface defined by $\mathbf{x}(\mathbf{p}) = (u, v, z(\mathbf{p}))$, the normal is derived from its partial derivatives:
\begin{equation}
\mathbf{n}(\mathbf{p}) \propto \left(-\frac{\partial z}{\partial u}, -\frac{\partial z}{\partial v}, 1\right).
\label{eq:normal_from_depth}
\end{equation}
By optimizing over $z(\mathbf{p})$ alone, this approach ensures that the recovered depth and normals are always geometrically consistent, producing a valid, integrable surface.

For stable and efficient optimization, we also follow \cite{NLiPs} and \cite{prados2003perspective} to introduce a log-depth parameterization. This is achieved by setting our optimization variable to be $\tilde{z}(\mathbf{p}) = \log z(\mathbf{p})$. This transformation offers two key advantages: it removes the physical constraint of positive depth ($z>0$) by allowing the unconstrained variable $\tilde{z}(\mathbf{p})$ to span all real numbers, and it helps improve numerical stability by linearizing the image formation model.

With this re-parameterization, the calibration problem can be reformulated as minimizing a global energy functional \(E(\tilde{z}, \tilde{\rho})\), which jointly estimates the log-depth map \(\tilde{z}\) and the ``effective albedo'' \(\tilde{\rho}\) from the observed images:

\begin{figure}[h]
    \centering
    \includegraphics[width=0.95\linewidth]{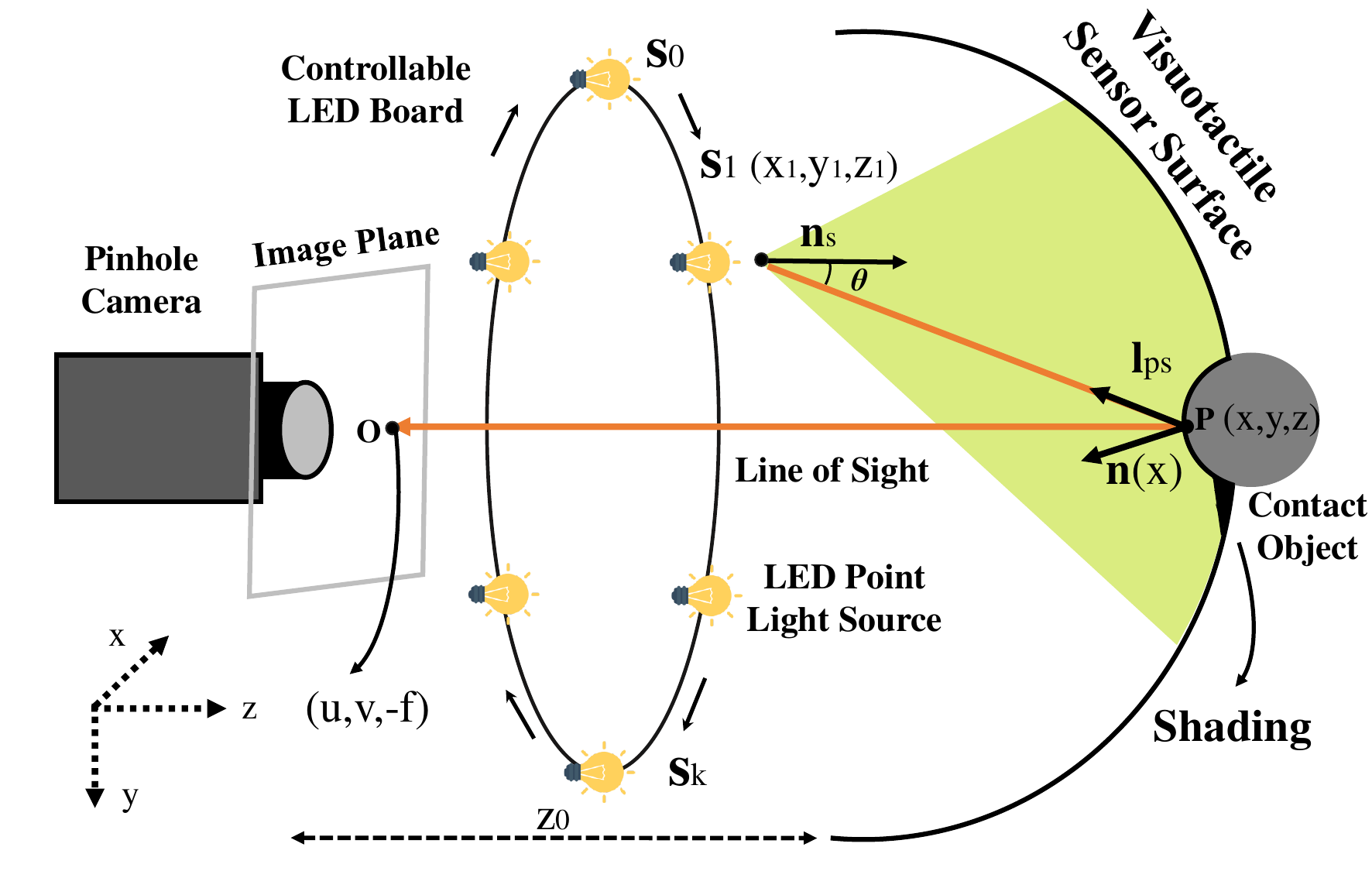}

    \caption{Optical path of near-light photometric stereo calibration, showing illumination from embedded point light sources to the sensor surface, which is then observed by the camera. Controllable LED point light sources are arranged in a ring ($S_0\!-\!S_k$) with known coordinates, and illumination from a selected source (e.g., $S_1$) defines the shaded region.}

    \label{fig:Lighting Modeling}
    \vspace{-8mm}
\end{figure}

\begin{align}
\min_{\tilde{z}, \tilde{\rho}} \; E(\tilde{z}, \tilde{\rho}) 
&= \sum_{\mathbf{p},i,c} 
\Big( I^{\text{obs}}_{i,c}(\mathbf{p}) 
   - I^{\text{pre}}_{i,c}(\mathbf{p}; \tilde{z}, \tilde{\rho}) \Big)^2 \notag \\
&\quad + \zeta \sum_p \big( \tilde z(p) - \tilde z_0(p) \big)^2,
\label{eq:optimization_energy}
\end{align}
where the first term penalizes the squared difference between the observed intensities ($I^{\text{obs}}$) and those predicted by our reformulated model ($I^{\text{pre}}$), and the second term is a regularization term with respect to an initial prior $\tilde z_0$, whose strength is controlled by the weight $\zeta$. {Following \cite{NLiPs}, the prior $\tilde{z}_{0}$ is initialized using the nominal camera--surface distance. Although this initialization influences the optimization process, we found the reconstructed local shape is robust to variations of $\tilde{z}_{0}$. }

To solve this optimization problem, we follow \cite{NLiPs} and use an Alternating Reweighted Least Squares (ARLS) algorithm. This method iteratively refines the solution by alternating between two main steps:
\begin{enumerate}
    \item \textbf{Albedo Update:} With the current log-depth map $\tilde{z}$ held fixed, the problem of finding the optimal effective albedo $\tilde{\rho}_c$ simplifies to a pixel-wise linear least-squares problem, which has a closed-form solution.
    \item \textbf{Depth Update:} With the albedo $\tilde{\rho}_c$ fixed, the energy functional is linearized with respect to the log-depth $\tilde{z}$. This results in a large but sparse system of linear equations, which is solved efficiently for an updated $\tilde{z}$ using a Gauss-Newton step with a preconditioned conjugate gradient (PCG) solver.
\end{enumerate}
By iteratively applying these two steps, the energy functional decreases monotonically until convergence. {\color{red}}The final output is a high-fidelity logarithmic depth map $\tilde{z}$, from which we recover the physical 3D surface $z = \exp(\tilde{z})$ and its corresponding normal field, serving as the ground-truth geometry for our calibration dataset.

\begin{figure*}[h]
    \centering
    \includegraphics[width=0.96\linewidth]{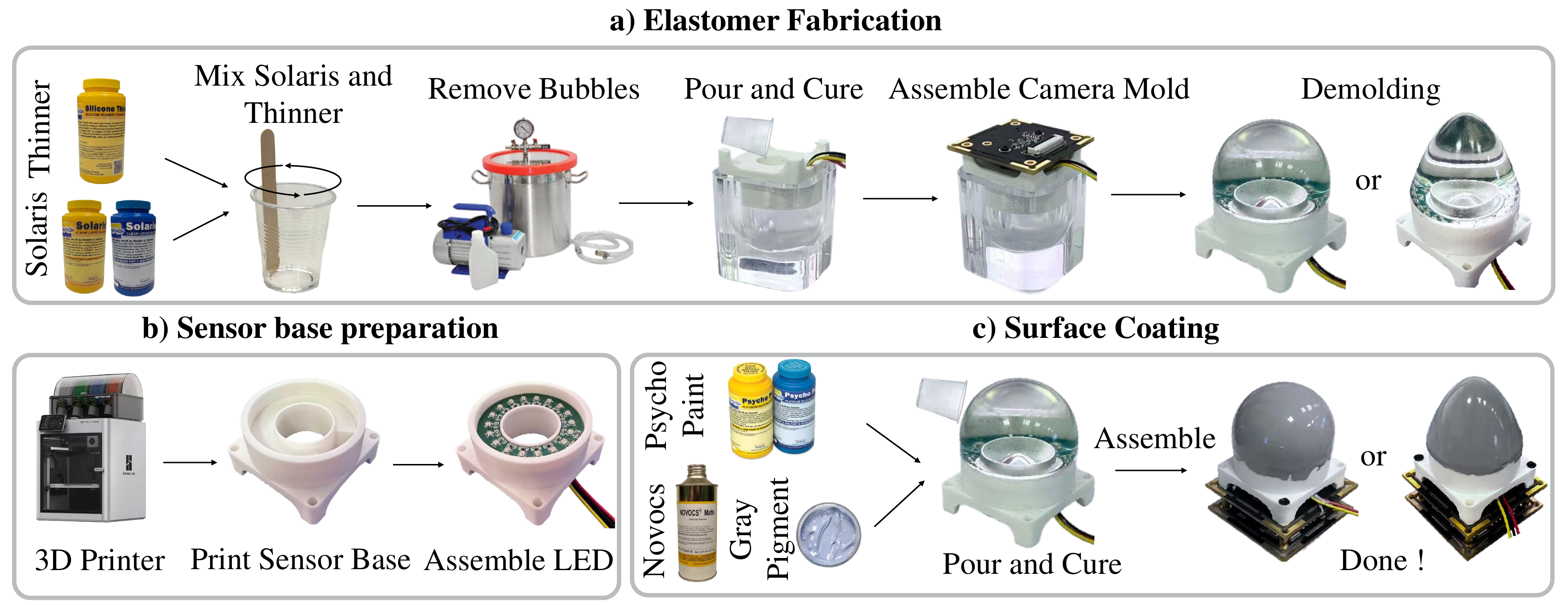}
    \caption{Fabrication of the NLiPsTac sensor, including (a) casting a clear elastomer for light transmission and contact support, (b) assembling the 3D-printed base with the LED board, and (c) casting a coating layer to form the reflective surface.}
    \label{fig:fabrication}
    \vspace{-4mm}
\end{figure*}

\subsection{Real-Time Normal Estimation} \label{sec:real_time_normal_network}

Near-light photometric stereo excels in reconstruction quality but is not well suited for real-time applications. This limitation arises primarily because it requires capturing multiple images, which increases acquisition time, and because the optimization process is computationally expensive.
To address this issue, we first construct a training dataset using the approach introduced in Sec.~\ref{sec:calibration_method}, and then develop a neural network to accelerate the inference process.

\paragraph{Dataset Acquisition} \label{dataset acquisition}

The calibration dataset is collected through indentation experiments. Unlike previous works, our approach (Sec.~\ref{sec:calibration_method}) allows direct indentation of the sensor’s elastomer using textured everyday objects. 

For each indentation, we record a set of calibration images. Using a sensor with 12 LEDs as an example, we collect: (i) 12 single-light RGB images, each with one LED activated under \emph{white} illumination; (ii) one \emph{dark} image with all LEDs off; and (iii) one tri-chromatic RGB image with all 12 LEDs activated simultaneously in three color groups (1–4 red, 5–8 green, 9–12 blue). The NLiPs model uses the single-light and dark images to reconstruct surface normals of the deformed region, which serve as ground-truth supervision. This enables the network to learn a mapping from tri-chromatic RGB images to surface normal ground truths.

\paragraph{Network Architecture}

We then propose \textit{NLiPsNet}, which infers the normal vector $\mathbf{n}$ from a 5-dimensional input $(u, v, r, g, b)$ (Eq.~(\ref{eq:normal_principle})). The network takes as input a single tactile image captured with all LEDs activated, following the color configuration in \cite{gomes2020geltip}.

NLiPsNet is implemented as a lightweight multilayer perceptron with three hidden layers (256–256–128) and an output layer predicting a 3-dimensional normal vector. Batch normalization, ReLU activation, and dropout (p = 0.2) are applied to improve training stability and reduce overfitting. During both training and inference, the predicted normals are normalized to unit length.

\paragraph{Training Objective}
We use a cosine similarity loss to train the network, enforcing angular alignment between the predicted normal $\hat{\mathbf{n}}_i$ and the ground-truth normal $\mathbf{n}_i$ for each valid pixel $i \in M$:
\begin{equation}
\mathcal{L}_{\mathrm{cos}} = \frac{1}{|M|}\sum_{i\in M}\left(1-\frac{\hat{\mathbf{n}}_i\cdot \mathbf{n}_i}{\|\hat{\mathbf{n}}_i\|\|\mathbf{n}_i\|}\right).
\end{equation}


\vspace{1mm}
\section{Sensor Design and Fabrication}
\label{sec:Sensor Design}

\subsection{Design Goals} \label{sec:normal_estimation}
We present NLiPsTac, a visuotactile sensor designed to validate our NLiPsCalib framework. NLiPsTac includes three key features: (i) integration of multiple individually controllable LEDs, providing directional illumination of the coating; (ii) does not include a light diffuser. Unlike sensors that approximate parallel lighting, each LED in our design acts as a point light source; and (iii) housing both the camera and LEDs within a single optical medium, minimizing refraction effects. These design choices align the sensor’s optical behavior with the physical assumptions of the NLiPs model, reducing model mismatch and simplifying algorithmic adaptation. The fabrication process of NLiPsTac is shown in Fig.~\ref{fig:fabrication}.

\subsection{Details of Sensor Design and Fabrication}
\textbf{Camera.} As the core imaging unit of the visuotactile sensor, we use an IMX274 camera module (30 FPS, Sony Inc.) equipped with an M14 lens. The 8 MP CMOS sensor supports a streaming resolution of 4K (3840$\times$2160 pixels). M14 optical lens provides a 100° field of view for comprehensive visualization of the dome-shaped elastomer surface.

\textbf{Sensor Base.} The base is a 3D-printed PLA structure that supports the LED circuit board, elastomer, and the camera module.

\textbf{LED Circuit Board.} The LED circuit board provides illumination for the sensor. It is a printed circuit board (FR4 substrate) populated with WS2812 addressable LEDs. Since near-field photometric stereo requires illumination from multiple directions, each LED can be individually controlled to emit trichromatic light at a user-specified intensity.

\textbf{Elastomer.} The elastomer is designed to transmit light to the surface, support the reflective coating, and withstand mechanical contact during deformation. Following \cite{vitrani2025shadowtac}, we used Solaris (Smooth-On Inc.) as the base gel material and incorporated Thinner (Smooth-On Inc.) to enhance deformability by reducing hardness. The gel casting process employed a custom acrylic mold to avoid surface mold lines. The A and B components of Solaris were mixed with Thinner in a 1:1:0.8 ratio, and the sensor base was secured in the mold before pouring. Following \cite{wu2025humanft}, a camera mold was carefully inserted into the reserved hole to ensure a seamless fit between the camera and elastomer.

\textbf{Reflective Coating.} The reflective coating is designed to approximate Lambertian reflectance. For fabrication, we used Psycho Paint and NOVOCS solvent (Smooth-On Inc.) with gray pigment. Following  \cite{lin2025pp}, the A and B components of Psycho Paint, NOVOCS solvent, and gray pigment were mixed in a 1:1:1:0.3 ratio, and the resulting solution was uniformly sprayed onto the demolded elastomer surface. 

\vspace{-1mm}
\section{Experiments}
In this section, we conduct experiments to validate the proposed calibration method, aiming to answer the following questions: 
(Q1) Can the NLiPsCalib approach obtain high-fidelity calibration results (Sec.\ref{sec:exp_calibration_data})?
(Q2) Can NLiPsNet achieve real-time, high-fidelity normal estimation (Sec.\ref{sec:exp_RTR})? (Q3) Can the calibration pipeline be applied to elastomers of different form factors (Sec.\ref{sec:exp_generalization})? In addition, we perform extensive experiments to investigate factors that influence calibration fidelity (Sec.\ref{sec:exp_ABT}).

\vspace{-2mm}

\subsection{Validation of NLiPs-Based Calibration}\label{sec:exp_calibration_data}

\begin{figure}[t]
    \centering
    \includegraphics[width=1\linewidth]{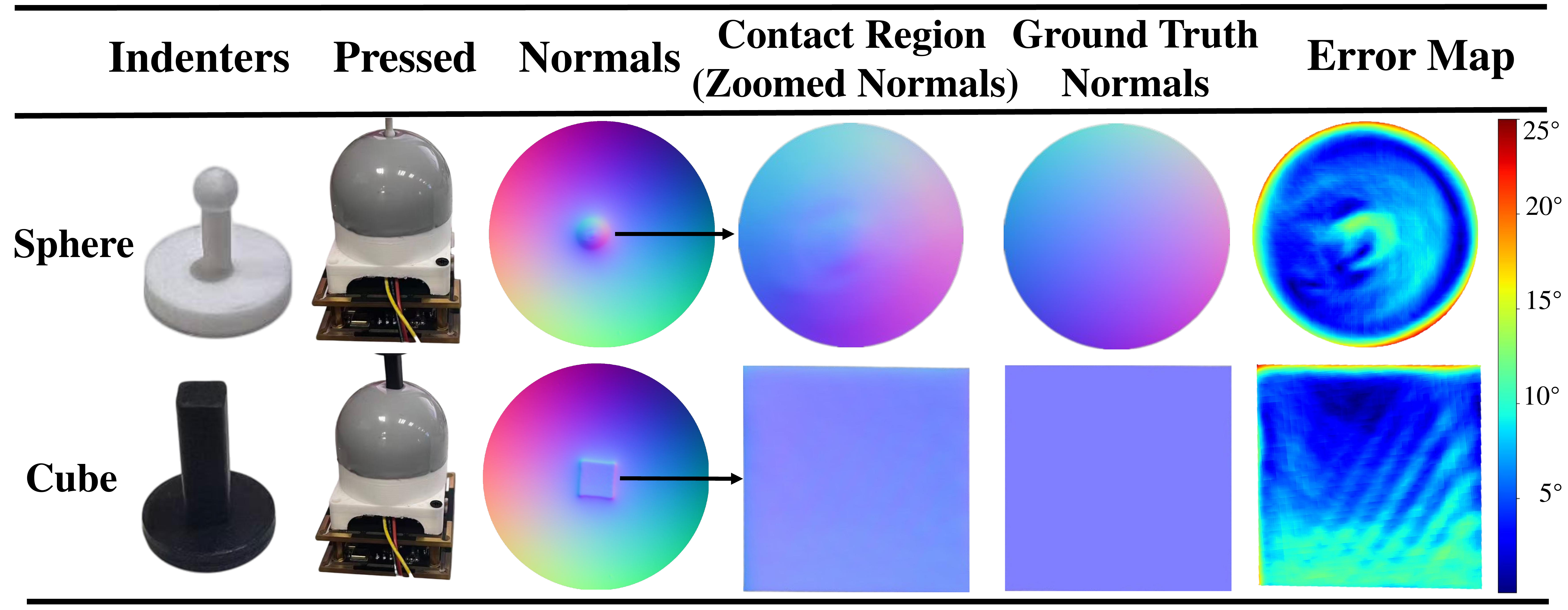}
    \caption{Normal estimation results for sphere and cube indenters. Columns show \textbf{Indenter}, \textbf{Pressed}, \textbf{Normals}, \textbf{GT (zoomed)}, \textbf{Normals (zoomed)}, and \textbf{Error Map}. The shared colorbar maps blue $\rightarrow$ red to errors from $0^\circ$ to $25^\circ$.}
  \label{fig:indenter_grid}
  \vspace{-6mm}
\end{figure}

We first evaluate whether NLiPsCalib can produce high-fidelity calibration results (Q1). 
To this end, we consider two representative cases: a small sphere (curved surface) and a cube (planar surface), both with analytically known surface normals. Using NLiPsCalib, the reconstructed normals were compared against the analytical ground truth. As shown in Fig.~\ref{fig:indenter_grid}, the calibrations closely match the analytical normals across the sensor.

For quantitative evaluation, we report accuracy using the Average Angular Error (AAE, ranging from $0^\circ$ to $180^\circ$), which measures the angular deviation between reconstructed and ground-truth normals, and the Mean Absolute Error of normal components (MabsE), which measures the absolute difference across the $x$, $y$, and $z$ components:\(\frac{1}{3|M|}\sum_{i \in M} \|\hat{\mathbf n}_i - \mathbf n_i\|_1\).

Our method achieves an AAE of $7.0415^\circ$ and an MabsE of $0.0588$. 
These results confirm that NLiPsCalib produces high-fidelity calibration results, validating its effectiveness for sensor calibration (Q1).

\vspace{-2mm}

\subsection{Validation of Real-Time Normal Inference}\label{sec:exp_RTR}

\begin{figure}[t]
    \centering
    \includegraphics[width=1\linewidth]{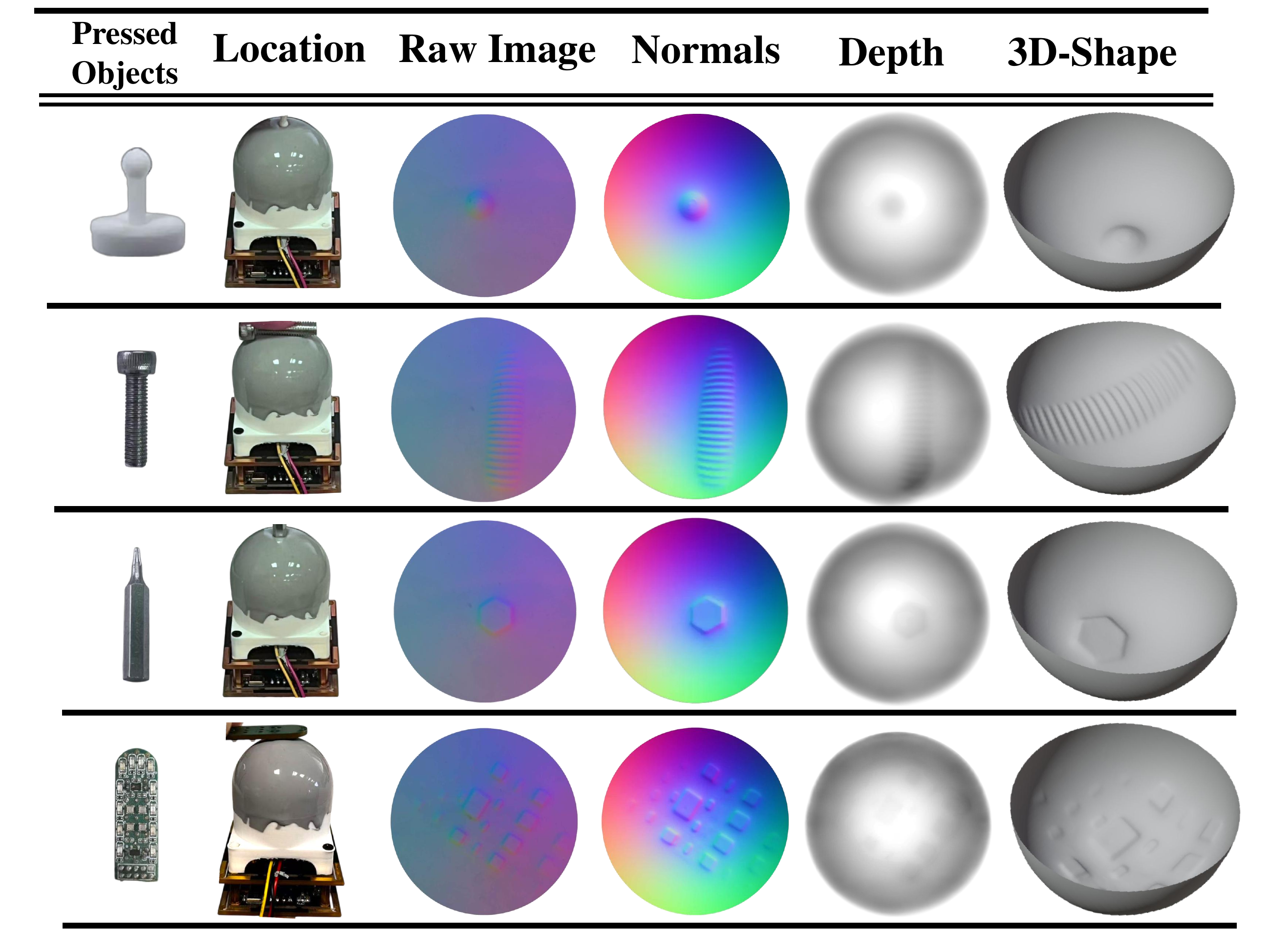}
    \caption{Samples from the NLiPsNet training dataset. Each row shows one pressing instance: the input \emph{probe} image, \emph{indentation location}, NLiPs-recovered \emph{normals}, and  \emph{3D shape}.}
    \label{fig:calib_dataset_grid}
    \vspace{-6mm}
\end{figure}

\textbf{Dataset Preparation.} A training dataset is required to train the normal inference network NLiPsNet. To prepare this dataset, we press three distinct objects at multiple sensor locations, resulting in \textbf{50} pressing conditions following the approach described in Sec.~\ref{sec:real_time_normal_network} (a). Some samples from the collected dataset are shown in Fig.~\ref{fig:calib_dataset_grid}.

\begin{figure}[t]
    \centering
    \includegraphics[width=1\linewidth]{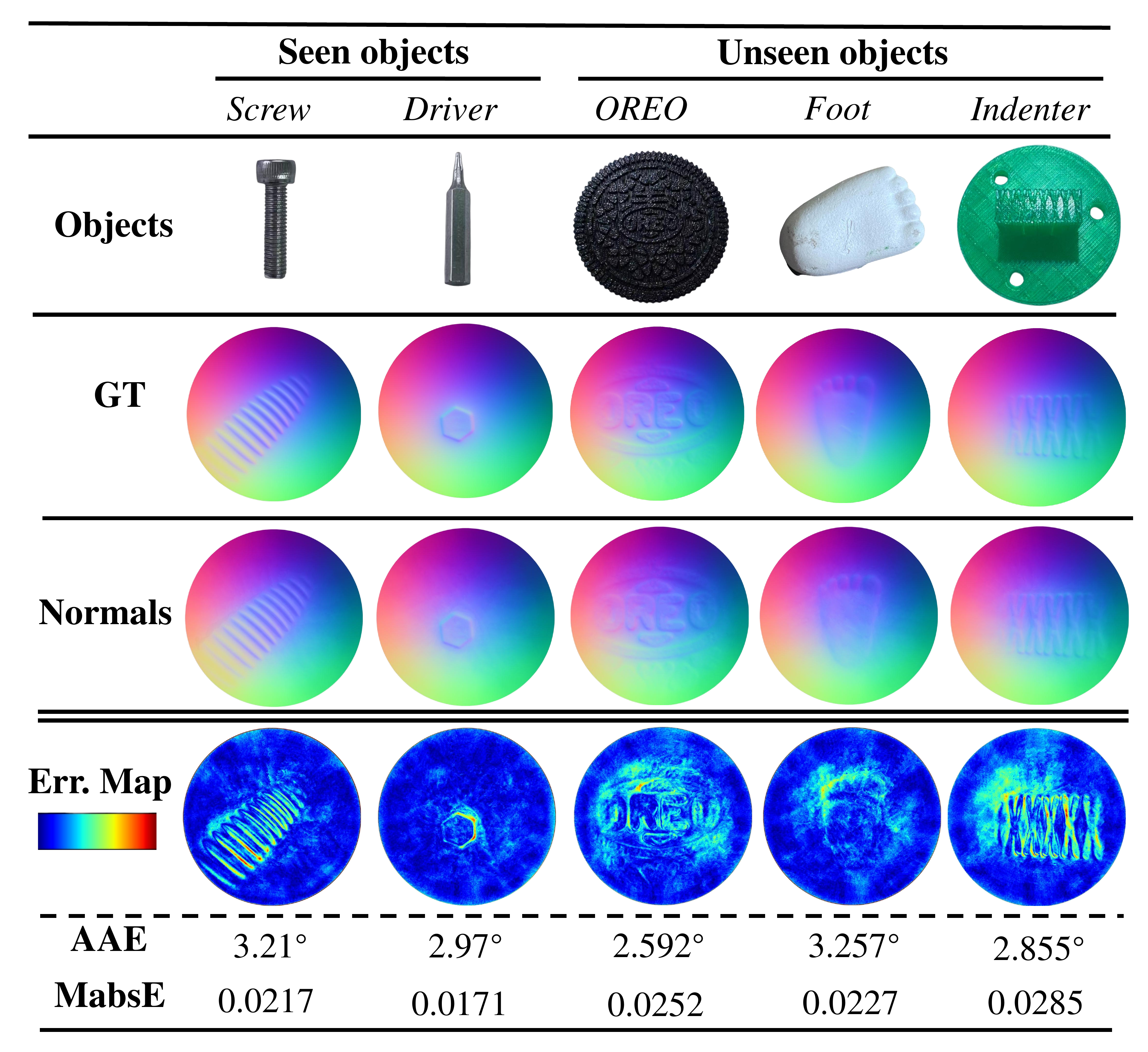}
    \caption{Evaluation results of NLiPsNet. Columns 1–2 correspond to objects seen during training, and Columns 3–5 to unseen objects. Rows show the pressed \textbf{Objects}, \textbf{GT} from NLiPsCalib, predicted \textbf{Normals}, and the corresponding \textbf{Err. Map} with ground truth from NLiPsCalib. The colorbar (blue to red) indicates errors from 0.00 to 0.05 in MabsE. }
\label{fig:compare_normals}
\vspace{-6mm}
\end{figure}

\textbf{Normal Inference.} We next evaluate the real-time normal inference capability of NLiPsNet. To assess both accuracy and generalization, we test the network using (1) indenters included in the training dataset but applied at unseen locations, and (2) entirely novel indenters. For all test cases, we also perform NLiPs-based reconstruction to obtain ground-truth normals, enabling both qualitative and quantitative comparisons. Example indentation observations, predicted normals, and NLiPs-derived ground truth are shown in Fig.~\ref{fig:compare_normals}. The predicted normal maps closely match the ground truth across all cases, demonstrating that NLiPsNet generalizes well to novel contact geometries.

Quantitative results (Fig.\ref{fig:compare_normals}) show that the network achieves an average angular error of $3.332^\circ$ on training-set indenters and $3.113^\circ$ on unseen indenters. These results highlight the strong generalization capability of NLiPsNet. The error levels are comparable to or better than many recent visuotactile sensors, which often report higher errors (e.g., GelRoller\cite{zhang2024gelroller} reports $16.17^\circ$). This validates that NLiPsCalib can effectively generate training datasets using only a few everyday objects and a few dozen indentations (Q2).

\begin{figure}[t]
    \centering
    \includegraphics[width=1\linewidth]{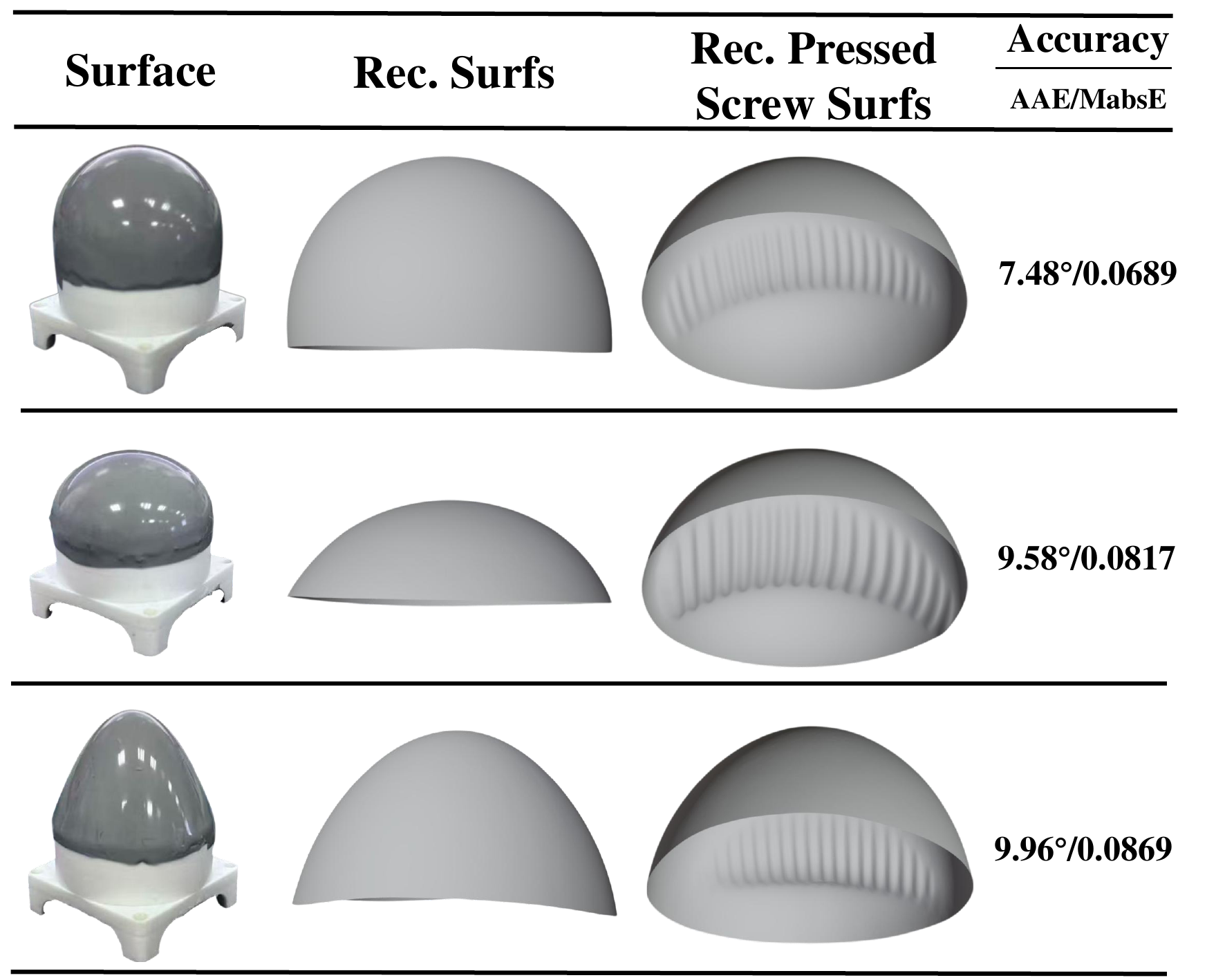}
    \caption{The proposed approach is applicable to diverse curved elastomer surfaces. Each row corresponds to a different elastomer dome geometry. Columns show the fabricated \textbf{Surface}, reconstructed \textbf{Rec. Surfs}, reconstructed \textbf{Pressed Screws}, and quantitative \textbf{Accuracy} (AAE, MabsE).}
    \label{fig:surface_recon}
\end{figure}

\vspace{-2mm}

\subsection{Generalization Across Curved Elastomer Surfaces}\label{sec:exp_generalization}  

We further demonstrate that the proposed techniques can be applied to visuotactile sensors with diverse curved geometries. For this purpose, we designed and fabricated elastomer domes with three distinct surface shapes. For each shape, we used NLiPsCalib to reconstruct the geometry and compared it against the corresponding ground-truth model, as shown in Fig.~\ref{fig:surface_recon}. Following the evaluation protocol in Sec.~\ref{sec:exp_calibration_data}, we quantitatively measured calibration accuracy for each case. 
The results show that our approach consistently achieves less than 10° AAE calibration error while preserving fine-scale details across varying elastomer geometries. These findings confirm that the proposed calibration method maintains high accuracy and robustness regardless of the underlying elastomer curvature (Q3).

\vspace{-3mm}

\subsection{Influence of LED Channels}\label{sec:exp_ABT}
\vspace{-1mm}
Since our approach leverages highly redundant LED illumination, we conducted an ablation study to analyze the impact of the number of LED channels on NLiPsCalib performance. We varied the number of active LEDs during calibration, testing configurations from 3 to 24 LEDs, and evaluated calibration accuracy by comparing the recovered geometry of a spherical indenter against the ground-truth shape. The results are summarized in Tab.~\ref{tab:ablation_leds}.

Results indicated that calibration accuracy depends on the number of LEDs. While acceptable results can be obtained with as few as 3 LEDs, increasing to 12 LEDs significantly improves both AAE and MabsE, indicating that denser illumination provides richer cues for surface normal estimation. Beyond 12 LEDs, however, improvements plateau: using 12 LEDs already yields a relatively low AAE, and 18 or 24 LEDs only slightly reduce MabsE. These results suggest that additional LEDs offer marginal benefits. Considering that more LEDs linearly increase calibration time with limited accuracy gains, we adopt 12 LEDs in our main experiments as a practical trade-off.

\begin{table}[t]

\centering
\renewcommand{\arraystretch}{1.1}
\setlength{\tabcolsep}{3pt}
\caption{Performance of NLiPsCalib with 3–24 active LEDs used for calibration and reconstruction.}
\label{tab:ablation_leds}
\resizebox{0.98\columnwidth}{!}{%
\begin{tabular}{@{}c!{\vrule width 0.5pt}cccccccc@{}}
\toprule
\#LEDs & 3 & 6 & 9 & \textbf{12} & 15 & 18 & 21 & 24 \\
\midrule
AAE $\downarrow$   & 9.86 & 8.48 & 8.17 & \textbf{7.28} & 7.84 & 7.17 & 7.81 & 7.46 \\
MabsE $\downarrow$ & 0.0858 & 0.0754 & 0.072 & \textbf{0.065} & 0.069 & 0.065 & 0.069 & 0.063 \\
\bottomrule
\end{tabular}%
}
  \vspace{-6mm}
\end{table}

\vspace{-1mm}
\section{Conclusion}

We presented NLiPsCalib, an easy-to-use framework for calibrating curved visuotactile sensors. Our approach addresses a key bottleneck in the development of custom tactile sensors: the dependence on costly devices, and labor-intensive data acquisition procedures. Leveraging a NLiPs model, NLiPsCalib generates high-fidelity ground-truth geometry directly from the sensor’s internal illumination, requiring only a few simple contacts with everyday objects and eliminating the need for specialized hardware. Experiments on the NLiPsTac sensor validate the effectiveness of this approach, showing that NLiPsCalib achieves calibration efficiency and accuracy comparable to state-of-the-art methods while significantly reducing setup time and complexity. The resulting calibration data enables the lightweight neural network NLiPsNet to perform accurate, real-time 3D reconstruction from a single image, demonstrating strong generalization to unseen objects and contact geometries.

A current limitation is the computational cost of NLiPs optimization during offline calibration (3-4 minutes per reconstruction, and 3 hours for overall calibration without human intervention), as our implementation is CPU-based and not parallelized. Future work will focus on improving computational efficiency of NLiPs, potentially through GPU acceleration. Moreover, the selection of suitable objects for calibration is also still empirical. Although the quantitative impact is difficult to measure, we provide the following guidance: objects should fully excite the sensor’s normal patterns. To capture diverse normal conditions, we recommend using a few different objects that produce salient deformations and pressing them across locations covering the entire elastomer. In addition, when LEDs illuminate the surface in sequence, the object should remain stationary.

{\small
\bibliographystyle{IEEEtran}
\bibliography{reference}
}


\end{document}